# Reprogrammable Surfaces Through Star Graph Metamaterials

Sawyer Thomas[1], Jeffrey Lipton[1]*

## ABSTRACT

The ability to change a surface's profile allows biological systems to effectively manipulate and blend into their surroundings. Current surface morphing techniques rely either on having a small number of fixed states or on directly driving the entire system. We discovered a subset of scale-independent auxetic metamaterials have a state trajectory with a star-graph structure. At the central node, small nudges can move the material between trajectories, allowing us to locally shift Poisson's ratio, causing the material to take on different shapes under loading. While the number of possible shapes grows exponentially with the size of the material, the probability of finding one at random is vanishingly small. By actively guiding the material through the node points, we produce a reprogrammable surface that does not require inputs to maintain shape and can display arbitrary 2D information and take on complex 3D shapes. Our work opens new opportunities in micro devices, tactile displays, manufacturing, and robotic systems.

## MAIN TEXT

In the natural world, rapid shifts in texture allows animals such as frogs, cuttlefish, and octopi to blend into their surroundings[1–3], and in our own fingertips, the wrinkling of skin improves our ability to grip objects underwater[4]. Current morphing surfaces rely either on individual actuators to drive and maintain system states, making them large, inefficient, and difficult to build[5–8], or they must be programmed at construction, limiting the range of accessible states for a single structure[9–17]. Despite the utility found by nature, the ability to produce varied and passively stable surface changes in mechanical devices on demand has eluded us. We have developed reprogrammable metamaterials that can generate arbitrary surface profiles and be rewritten after fabrication to utilize kinematic singularities and a star-graph state trajectory structure. These metamaterials have a transition state where small inputs can cause the system to have significantly different responses to global loading. We use this to program both global and spatially varying Poisson's ratio, program edge profiles, display 2d information and make fully developable surfaces. Our results demonstrate how the structure of the state space of a mechanical metamaterial can be used produce writable and stable constructs that change shape and mechanical properties.

## BACKGROUND

Mechanical metamaterials can demonstrate unusual properties based on their architected periodic structure[18–21]. They are often represented as a graph of links that are embedded in a space[22,23]. This space can be finite as in a sphere[24], closed like a cylinder[25], or infinite and open as in a plane[26,27] or 3D space[15,22,26]. The embedding of the link graph reduces the infinite degrees of freedom of a plane or space filled with links into a small finite number of degrees of freedom. In the case of auxetic metamaterials, the number of degrees of freedom reduce to a single degree of freedom θ, often an angle between two links, that defines the evolution of the system[25]. This θ defines a trajectory of states, where the system must kinematically evolve by either increasing or decreasing θ, and the range of θ determines the limits of the system's deformation[25].

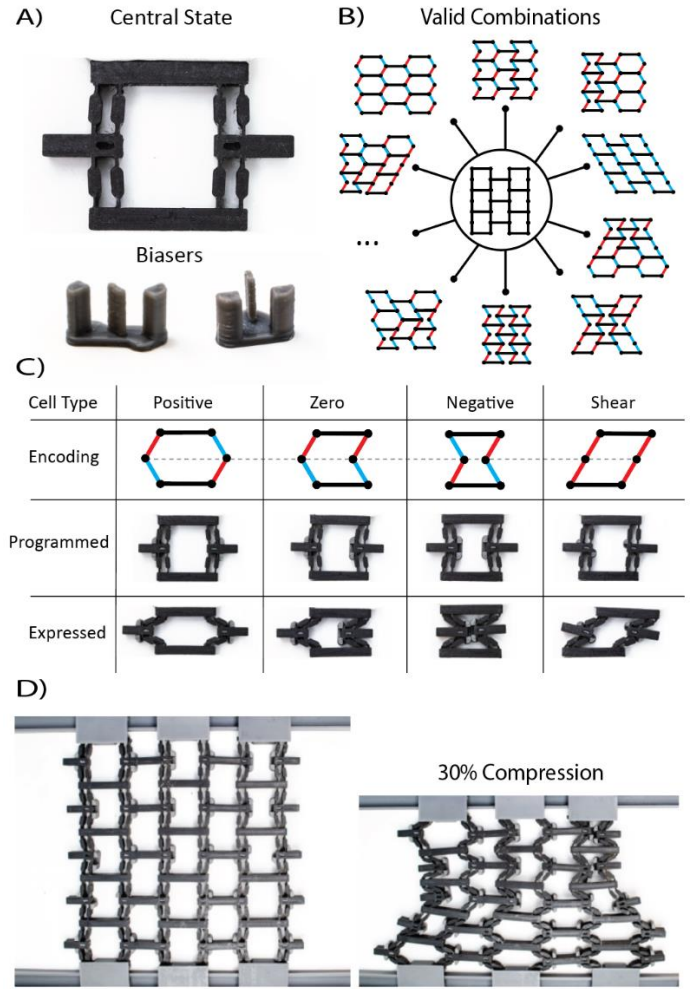

**Figure 1 | Reprogrammable Star Graph Configuration. A)** A single compliant 4-bar unit cell (top) and 3D printed blockers for constraining the deformation of the cell (bottom). **B)** An $A \times B$ lattice with $A = 6$ and $B = 4$ vertical linkages can transition into 6642 different valid configurations, matching a star graph structure (SM). **C)** A single $2 \times 2$ linkage unit cell can be reprogrammed to 4 cell types (6 including mirrors), each with different mechanical properties. **D)** Encoding different unit cells throughout a structure enables complex deformations of a 3D printed lattice.

[1] Department of Mechanical Engineering, University of Washington, Seattle WA USA, 98195

To enable programable surface changes, we developed metamaterial structures that have multiple trajectories, each representing a different surface profile, and provide a means for moving between these kinematic trajectories. The key concept for enforcing transitions between these trajectories is physically constraining mechanical singularities (Fig.1). Mechanical singularities are often viewed as terrible states for a system and normally should be avoided. Many quantities become infinite or indeterminate at these points. The canonical example is gimbal lock, where two Euler angles become degenerate, and it is impossible to disentangle the two angles. In robotics a singularity in a joint may require infinite joint velocities to maintain a smooth movement or could generate infinite inverse-kinematic solutions.

**RESULTS**

We present a subset of lattices that can be actively controlled to morph between valid physical states by leveraging this property of mechanical singularities. For certain materials, there are a series of different state trajectories that represent valid embeddings of the link graph. Each of these trajectories has a single degree of freedom $\theta$, that is typically an angle between links. Each possible physical embedding is a node in a star graph network with a single state acting as the central node (Fig.1.B). At this central node, all trajectories converge to a single point characterized by a mechanical singularity.

At this singularity, an N by M tiling of cells goes from a single degree of freedom to Nx(M+1) degree of freedom system (SM). The spontaneous emergence of these degrees of freedom is what makes this point unpredictable. The singular point is metastable[28] so small nudges can adjust the angles between links, transitioning the system between state trajectories. These small local nudges effectively program the path of the system (Fig.1.C). As you move away from the singular point, the structure develops multi-stability, where several low energy states exist with a high energy barrier separating them. Global compression guides the system along a selected path to reach an expressed state (Fig.1.C).

The number of valid embeddings with a single degree of freedom is limited. Many combinations will result in geometric restrictions, causing the system to become frustrated[10,29]. For a given combination of joint orientations to be valid, link lengths must be preserved throughout the system as $\theta$ evolves. A configuration can be validated by analyzing horizontal joint displacements at maximum compression to ensure that horizontal link lengths are preserved (SM).

Different physical embeddings of a link graph structure produce a bulk material response with different mechanical properties. As has been reported in the literature, the double arrowhead or the honeycomb patterns can achieve the extremal global Poisson's ratios of positive one half and negative one along with zero Poisson's ratio[20]. The global Poisson's ratios can be varied between +0.5 and -1 by establishing strips of the same cell type over rows of the lattice. The ratio of cell types governs the magnitude of Poisson's ratio. Therefore, as seen in Fig.1.B the system can achieve a discrete number of values and transition between them at the node state. The striping of various embeddings can also be used to produce preprogrammed profiles with a vase like nature[11,30].

Generating arbitrary profiles requires a cell to be capable of both positive and negative Poisson's ratios and to have a shearing state. The shear element (Fig.1.C) represents a completely different response that is not defined by Poisson's ratio. Therefor a lattice must have at least one auxetic state trajectory, a transition point between auxetic an non-auxetic, and the ability to have a shear mode. We found the honeycomb patterns fits all the criteria for a programmable link structure. As demonstrated in Figure 1.D (Supplementary Video 1) the shear response can connect regions with positive and negative Poisson's ratios, enabling spatially varying shape changes throughout the material. If no shearing layers are included, attempting to alternate cell types between adjacent horizontal rows will result in discontinuities along the edge of the material, causing geometric frustration and failure. The shear cell is a necessary topological component to bridge the difference in cell width and generate complex profiles (Fig.2.A).

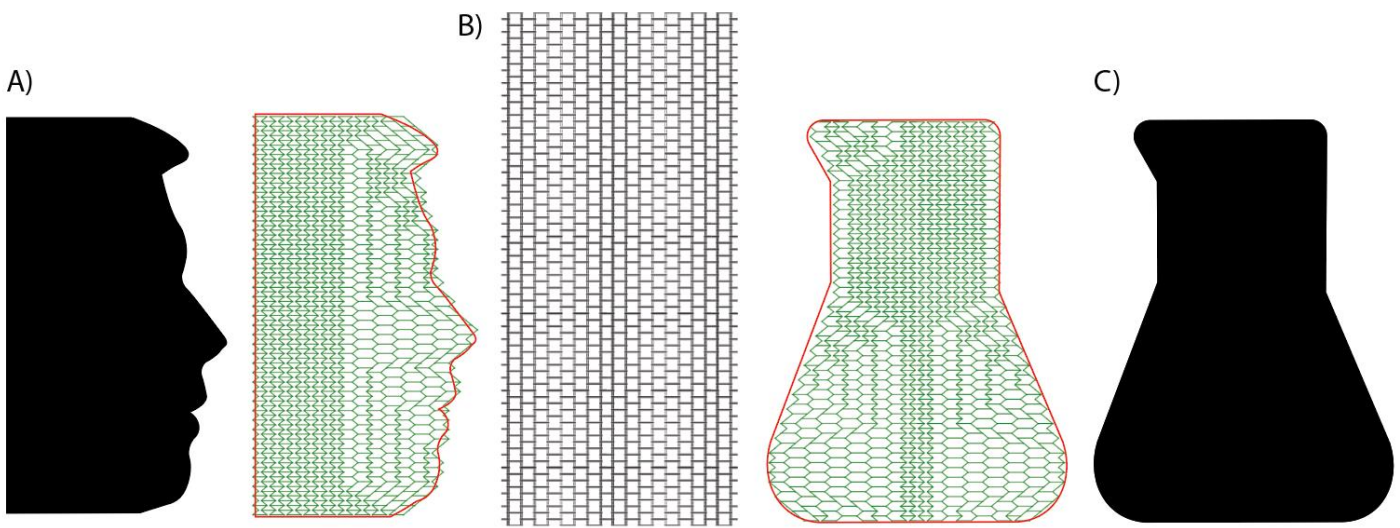

**Figure 2 | Complex Shape Generation** A) Algorithmic profile generation allows us to approximate a flat backed face profile by programming 17 × 41 cell lattice. B) The extended state of the lattice acts as the central node of a star graph with many possible programmable trajectories. C) The same 17 × 41 cell lattice is encoded to match the shape of a beaker with asymmetric profiles on each side.

For a honeycomb tiling with $A \times B$ vertical linkages, the total number of possible combinations is $2^{A \times B}$. Each vertical linkage is encoded as either a positive or negative value which corresponds to its slope as it compresses. To guarantee a frustration free trajectory[10], all linkage lengths must remain constant throughout the deformation of the system. To test the validity of any state, we created an algorithm to verify if an encoding maps to a valid configuration (Methods; SM). Using a brute force application of this validity test, we experimentally found the number of total valid states for all lattice combinations with $A * B$ less than 50. By fitting this data to the general equation $2^{k_1 AB + k_2(A+B) + k_3}$ we obtained $k_1 = 0.2989$, $k_2 = 0.6924$, $k_3 = -1.3831$ (Methods). Following this trend, the number of valid states grows quickly but the probability of randomly selecting a valid configuration from the transition state rapidly approaches zero as the size of the tiling grows. For a $10 \times 10$ array, there are approximately $5.62E12$ valid states but $1.2677E30$ possible combinations. Hence, the probability of selecting a valid state at random is only $4.433E - 16$ percent.

The sparsity of valid trajectories necessitates active control through the singularity point to ensure successful transitions from state to state. To do this we developed an algorithm to designate the state of each linkage in a lattice, based on a predefined edge profile (Methods; SM). Starting at the defined edge, columns are generated such that they match the conditions enforced by the validity test and approach the desired profile on the opposing edge (Methods; SM). This strategy allows us to generate complex profiles using large arrays to approximate detailed shapes (Fig.2).

To define the edge encoding of a high-resolution profile or function, we deconstruct the shape into a series of line segments (SM). The magnitude of the actuating compression determines the maximum slope of the vertical linkages. Combinations of multiple positive and negative elements allow edges to approximate intermediate slopes (Fig.2). These combinations of linkage slopes combine to generate complete profiles. With this model, we can construct high complexity profiles such as the silhouette of a face (Fig.2.A) or a beaker (Fig.2.C). Using the lattice encoding algorithm described above, we propagate this edge encoding backward through the structure to configure the full lattice shape.

To program these systems, vertical linkages must be mechanically biased to have either a positive or negative slope. We used multiple strategies to mechanically encode trajectories, including mechanical inserts (Fig.1.A) and small electromechanical actuators (SM). To demonstrate real-time reprogrammability, we constructed a lattice with distributed micro servos (SM). A linear axis, paired with a stepper motor, applied an axial compressive load to the lattice to transition the system into a specified compressed state. Encoding mechanisms constrain two adjacent linkages (above and below), meaning that biasers only need to be included at every other joint to fully program the system. Because the programming and the actuation of these structures is decoupled, the joint biasing forces and displacements can be extremely small (~.7 N) in comparison to the compressed holding force (~ 40 N). This opens a wide range of possible options for actively programming such as micro-actuators or shape-memory materials[3,6,12].

Combining multiple layers of these 2D lattice structures produces reprogrammable binary height fields (Fig.3). To generate a binary pattern, each planar lattice only requires a single layer of reprogrammable linkages. From the expanded state, the trajectory of each pixel will map to a height of either 0 or 1 based on the state of each edge linkage. A $R \times S$ grid, with $R$ lattice height and $S$ being number of layers, has a large design space with $2^{R \times S}$ valid combinations. Fig.3 shows a binary height field with several encodings. The displacement required to bias each joint is very small in comparison to the deformation expressed through global actuation. Because of this, the physically programmed state shows little variation between encodings while the information becomes clearly visible in the expressed state. This display can render any 6x7 binary pixel value, allowing us to render the entire English alphabet in block letters or brail (SM).

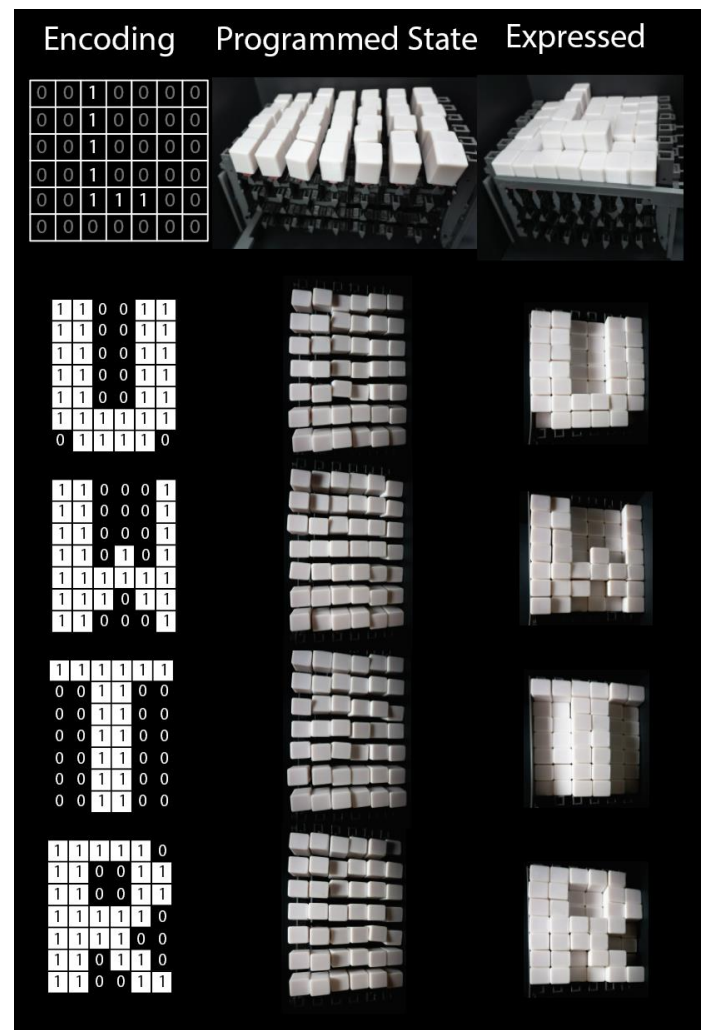

**Figure 3 | Binary Information Representation.** Multiple layers of the planar lattice structure are stacked, acting as a reprogrammable height map. Little variation can be seen between different physically programmed encodings. When we compress the structure, the expression of the encoding becomes apparent.

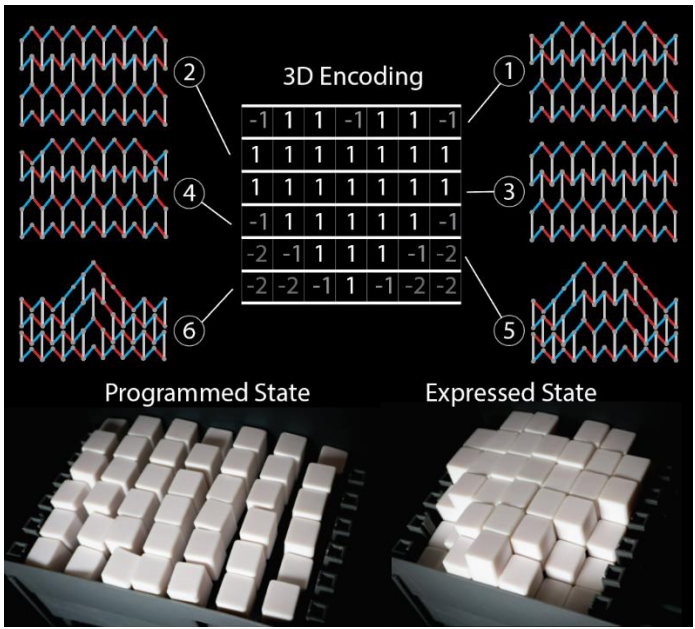

**Figure 4 | 3D Surface Expression.** Stacking multiple star graph lattice structures creates reprogrammable 3D height maps. A surface height encoding corresponds to layers of programmed 2D structures. The physically encoded information is expressed by compressing the structure.

We created reprogrammable 3D structures and surfaces by layering several planar structures in parallel. These structures can transform to generate both concave/convex curvatures and positive/negative space. Along the length of a layer, the individual height value between adjacent pixels cannot vary more than one unit step at a time. In between layers, no such restriction exists, and adjacent pixels can vary without encountering geometric frustration. Because of this relation, 3D profiles with large discontinuities in height can be approximated more accurately by aligning jumps in height with the inter-layer boundary. The main factors limiting surface profile representation are the depth/height of the design, the resolution of surface details, and the maximum slope of the design. Higher lattice cell counts would help to improve resolutions and slope can be governed by the cell geometry and the amount of compression.

## CONCLUSION

We have shown that metamaterials with star graph state trajectories can form the basis of reprogrammable surfaces. As demonstrated, specific combinations of programmable linkage angles within a lattice translate to mechanical shape changes in the structure. These structures have a convergent state of singularity, where all other valid states can be accessed. The compressive trajectories of these structures can be encoded using only small mechanical nudges to program the system. This approach supports a wide range of shape changes for both 2D and 3D structures. It decouples programming from actuation, creating opportunities for increased scalability and improved resolution. It also supports stable mechanical memory, needing no additional energy to hold a state once actuated. This concept is scale independent allowing for the strategy to work at the scale of MEMS devices up to architectural surfaces. Shape changing interfaces offer opportunities to fundamentally change human-computer interaction through object simulation, communication of visual and tactile information, user augmentation, and extended reusability[1,6]. Reprogrammable structures have utility in digitally adjustable tooling and jigs, variable friction materials, tunable acoustic surfaces[18], and robotic grippers, locomotion, and camouflage[16].

## METHODS

### Materials and Fabrication

To create both the honeycomb and the arrowhead structures (S7), we designed 3D printed lattices with compliant joints. We printed these from Ninjaflex Thermoplastic Polyurethane (TPU) using a Creality Ender3 printer. The lattices had a constant out of plane extrusion thickness of 6 mm. Rigid, vertical, and horizontal bars had a width of 3mm and a length of 20 mm. For the reentrant honeycomb, Individual cells formed approximate 28x28mm squares. We designed compliant joint components as standard knife blade flexures with a length of 2.25 mm and a width of .44 mm. To ensure an even compression of the structure, we replaced the single vertical linkages with two-bar linkages.

We manually reprogrammed the trajectory state of the lattices by inserting 3D printed blocking elements into the TPU lattice structure, effectively biasing the joints to buckle a specific direction (Fig.1.A). The blocking element deformed the joint approximately 1.3 mm and was rotated 180 degrees to switch the biasing angle from $\theta > \pi/2$ to $\theta < \pi/2$. An alternate blocking element locked the joint angle to be compatible with the shear cell configuration (Fig.1.A). We printed these blocking elements out of polylactic acid (PLA) using a Creality Ender3 printer.

To Make the 3D structures, we printed multiple 2D planar lattices and assembled supporting PLA components to make a layered rack system. To create a smooth surface when compressing the structure, we mounted 19x22x30 mm PLA caps on one edge of the mechanism (S8.B). We connected the edges of each TPU lattice to rigid rails using small PLA components with roller bearings that moved freely (S8.B) and maintained a 9.5 mm gap between sheets to avoid interference.

### Mechanical Tests

To ensure smooth compression, we fixed one cell on the top and bottom of the lattice to the compressing structure, establishing an origin in the lateral direction. We connected remaining cells to 10mm bearings to create a sliding connection between the cells and the rigid surfaces. We printed the compression structures out of PLA and actuated the system using a Nema 17 stepper motor and a motor mount position slide (McMaster Carr 6734K14). The samples rested on a Teflon sheet to reduce friction. We tested the biasing force for a fully expanded unit cell and the holding force for a compressed unit cell in both tension and compression using an Instron mechanical test setup (S12).

### Conditions for Valid Lattice Configurations

For a reentrant honeycomb lattice with AxB vertical linkages of length $H$, we developed criteria to test whether a given joint combination results in a valid state. At every leaf node of the star graph, each vertical linkage will have either a discrete positive or a negative slope. A 2D array of size AxB is populated with either 0s, denoting a negative slope, or 1s, denoting a positive slope (S4.A; S5.A.). Every possible valid state of the lattice makes up a finite subset within the total $2^{A*B}$ possible combinations of array values. Horizontal crossbars with a length of L alternate to connect every other grid point to the adjacent grid point, adding geometric constraints to the system. As the structure is compressed, all vertical linkages rotate to an angle $\pm\theta$, creating a horizontal offset of distance $a = H * \cos(\theta)$. Starting from the top-left edge of the lattice $(A, B = 0)$, we work across the top row and then down each column, populating a new array of size $(A + 1) \times B$ with each joint's horizontal offset. $Joint[0,0]$ is initialized to 0 and the remaining $joint[0,1:B]$ horizontal offsets are calculated such that, if n is odd,

$joint[0, n] = joint[0, j-1] + L$ and if n is even, $joint[0, j] = joint[0, j-1] + a_{i,j} + a_{i,j-1}$. Every other horizontal offset is calculated by stepping down each column $joint[1: A, B]$ and populating the array with $joint[i, j] = joint[i-1, j] + a_{i-1,j}$ (S4.B; S5.B). After calculating these distances, we can check the validity of the structure by verifying that the joint value to the right side of each horizontal crossbar is equal to the vertex value to the left side of the horizontal bar plus the width of the crossbar (L).

**Combinatorial Design Space**

To derive an expression for the total number of valid leaf nodes, we generated and tested every candidate combination of an AxB linkage array up to $A * B < 50$ and $A, B < 16$. This data created a symmetric matrix with 25 total points (S6). Of these combinations, we selected 17 points to act as fit data, and 8 points to act as validation data. By taking the $log_2$ of the fit data for valid configurations, we were able to generate best fit lines for $A = 2,4,6,8$ as B increased, with $R^2 > .99998$. The slope and intercepts of these four lines also fit a linear relationship as the A value increased, such that $R^2 > .99999$. This logarithmic relationship and the two linear equations combined to create a single general equation to describe the valid combination space as A and B varied. The number of total valid combinations $= 2^{k_1 AB + k_2(A+B) + k_3}$ with the three constants, $k_1 = 0.2989$, $k_2 = 0.6924$, $k_3 = -1.3831$, obtained through the linear fits. We tested this general equation using our validation data and achieved $error < 2.1\%$ for all points (S6).

**Flat Backed Lattice Configurations**

We created an algorithm to generate a valid lattice configuration for arbitrary edge conditions on one side of the lattice and a flat edge on the other side. Following the same notation as the combinatorial cell assigning algorithm, we represent each vertical linkage as a positive (1) or negative (0) value based on its slope. Starting with the defined profile, we step through each column of the lattice, reducing the difference between the maximum and minimum horizontal offset in the column ($joint_{max} - joint_{min}$). To fully configure each cell in the lattice we perform the state designating algorithm outlined in the body of the paper. For the flat back algorithm, instead of configuring cells to reducing the difference between opposite edges of the lattice, we aim to reduce the average horizontal offset for each column.

**Profile matching**

To match a compressed reentrant honeycomb lattice to an arbitrary profile, we first broke the profile into $n_s$ discrete line segments, each with a length $L_p$ and a slope angle $\theta_p$. Three factors limit the shape of profiles which we can accurately approximate. First, the distance between the minimum and maximum horizontal offset must be less than the maximum horizontal displacement of the lattice ($joint_{max} - joint_{min}$). Second, considering that $\theta_p = 0, \pi$ represents horizontal lines, the profile must be made up of line segments with slopes $0 < \theta_p < \pi$. This means that all profiles must be approximated as functions with a single x value mapping directly to a single y value. The value of $\theta_p$ is further limited by the dimensions of each cell (S1) since geometric interference will occur before the cell can be compressed completely flat. Finally, the total height of the profile must be less that the total height of the lattice structure in its compressed state. Here, the magnitude of compression can be governed either by the maximum slope $|\theta_p|$ or the minimum cell height required to accurately match the profile.

To approximate an arbitrary profile or function, we once again define the shape as a combination of positive sloped segments (1) and negative sloped segments (0). For a lattice with AxB linkages, we start by splitting a predefined profile into A segments and assigning each point an x and a y value. We then determine the distance that the lattice must compress so that each link height $(h)$ matches each profile line segment. Based on the magnitude of compression, we find $\theta_p = \arcsin(h)$ where $h$ is segment height. To fill in the lattice edge values, we start from the top and increment a horizontal displacement value. If the horizontal displacement value is greater than the current profile x value, then we include a 0 in our lattice edge array. Otherwise, we include a 1 in our array, repeating this process until the array is fully populated (S11; S15).

**Simulation of Large 2D profiles**

We performed simulations of the 2D profiles using Ansys static structural simulation tools. We set mesh size for the simulation to be a resolution of 7 and enabled large deformations. For structural constraints, we grounded the base of one cell at the bottom of each lattice and constrained the remaining bottom edge points, allowing deformation in only the x direction. We assigned negative 14.7 mm/cell z displacement at the edge of each top cell. All cell movement remained free in the x direction, except for one point on the top surface that we fixed, grounding the displacement.

We simulated the lattice in Fig.1.D to validate our lattice design strategies. To create a CAD lattice design, we manually combined a series of positively and negatively sloped joints in fusion 360. Rather than apply individual biasing forces at each joint, we assigned an initial slope offset of 5 degrees to each vertical linkage to establish the buckling direction. We simulated the global compression of the structure using the finite element approach described above (S7.B). To generate complex profiles such as those shown in Fig.2, we manually created profiles and then generated the corresponding lattice using our profile generation algorithm.

**Data availability**

The data supporting the findings of this study are available from the corresponding authors upon request. Source data are provided with this paper.

**Code availability**

The code used in this paper is freely available at GitHub

**Acknowledgements**

We thank the Murdock Charitable Trust, for their support through grant 201913596. The work was supported by the NSF through grants 2035717 and 2017927 and by the Office of Naval Research through grant DB2240. We would like to acknowledge Professor Lucas Meza for his assistance.

**Supplementary Materials:**

**Videos**

**Video 1:** The video *SV1.mp4* shows a 5 × 5 cell lattice structure, which is uniaxially compressed to 70% its original height using a Nema 17 stepper motor and a motor mount position slide. The structure deforms from the programmed state to the expressed state along the preprogrammed trajectory.

**Video 2:** The video *SV2.mp4* shows a 3 × 3 cell lattice structure, which is actively reprogrammed and uniaxially compressed to transition between 3 separate state trajectories. The lattice transforms between a negative Poisson's ratio structure, a positive Poisson's ratio structure, and a zero Poisson's ratio structure.

**Video 3:** The video *SV3.mp4* shows 4 separate trajectories of the 3D reprogrammable surface structure. The structure is manually reprogrammed and compressed. This video shows the transition between the programmed state and expressed state for a U, W, T, and R encoding (University of Washington, Transformative Robotics Lab).

**Lattice Selection**

To select a lattice with desired shape changing capabilities, three requirements must be met. First, the state space of the lattice must make up a star graph with all leaf nodes being accessible from a single central node. Second, the star graph configuration must have enough valid states to enable arbitrary shape change. For example, lattices such as the double arrowhead or chiral structures support the star graph configuration but have small state spaces that are limited by geometric constraints. To ensure that the number of valid states grows rapidly as cell count increases, adjacent cells must be independently programmable. Finally, individual cells of the structure must be capable of switching between discrete Poisson's ratios. For both the reentrant honeycomb and the double arrowhead structures, the Poisson's ratio can be set to either a positive or a negative value based on the interior joint angle of $\theta$. This property allows the width of compressed cells to be programmatically set, enabling shape change within the lattice.

Two common auxetic structures that can be dynamically switched between discrete positive and negative Poisson's ratios are the reentrant honeycomb[1–5] (S1.C) and the double arrowhead lattice[1,6–8] (S1.B). As shown in figure S1, the transition between positive and negative compressive trajectories occurs when each joint has an angle of $\theta = \pi/2$. As a result, any cell with an initial angle $\theta > \pi/2$ will expand laterally as the structure is compressed, and any cell with an initial angle $\theta < \pi/2$ will contract laterally as the structure is compressed. The rotating squares structure[9] (S1.D) is an example of a geometry that remains auxetic throughout the entire trajectory of $\theta$. Regardless of the initial bias of the expanded state, the structure will continue to compress laterally as it is compressed axially. This makes the rotating square structure a poor candidate for generating edge profiles.

To select a successful structure, we represent a complete lattice as a combination of mirrors and translations, with green dashed lines denoting mirrors in figure S1, and vectors $l_1$ and $l_2$ denoting translations. As $\theta$ develops between the angles 0 to $\pi$, the unit cell transforms continuously, scaling such that system matrix $\bar{G} = \begin{bmatrix} g_{11}(\theta) & 0 \\ 0 & g_{22}(\theta) \end{bmatrix}$. The instantaneous transformation of the system can be described as $\dot{\bar{L}}(\theta) = \bar{G}(\theta) \cdot \bar{L}(\theta)$, where matrix $\bar{L} = [l1, l2]$ defines the tiling of the unit cell and $\dot{\bar{L}}(\theta)$ is the instantaneous transformation of the tiling[9]. When the $\det(\bar{G}) = 0$ we can check the $Tr(\bar{G})$ to determine the trajectory of the system. If $Tr(\bar{G}) = 0$, then the system will remain either auxetic or non-auxetic throughout the range of $\theta$. If $Tr(\bar{G}) \neq 0$ then the system will switch from auxetic to non-auxetic at the point when $\det(\bar{G}) = 0$ [9]. Both the double arrowhead structure and the inverse honeycomb structure can be shown to exhibit this behavior. Star auxetic structures[10] also show this behavior but are not included within the scope of this paper.

**Star Graph Representations of Lattice Configurations**

Lattice structures such as those pictured in S2 and S3, can be represented as 2D kinematic linkages of rigid bars, transforming based on the angle $\theta$. If the value of $\theta$ is maintained as equal for every unit cell in the lattice, then the structure has a single trajectory path, and the state depends on one value of $\theta$ between $\theta_{min}$ and $\theta_{max}$. However, if the value of $\theta$ varies independently between separate unit cells, then the full lattice can transition between a larger set of different states. For such structures, a single state exists when $\theta = \pi/2$, such that every other state can be accessed by setting every individual unit cell angle to be $\theta > \pi/2$ or $\theta < \pi/2$. Once each initial value of $\theta$ has been set such that $\theta \neq \pi/2$, the compressive trajectory of the structure is fully defined, and the structure will continue to transition along a set path. This behavior can be represented with a star graph $S_n$ of order $n$ [11]. Here we can consider the internal node of the star graph to be the point at which $\theta = \pi/2$ for each cell. Every other accessible state of the lattice makes up the leaves of the star graph.

As shown in S2, the 4-bar chiral lattice[1] is one such geometry that can be described with this star graph data structure. Here, the current state of the structure can be defined by a combination of zero, clockwise, or counterclockwise center rotations and linkage rotations for each unit cell. As the cell count for the lattice increases, the chiral star graph has an exponentially increasing number of valid accessible leaves. However, each unit cell can only take on negative or zero Poisson's ratios and no positive Poisson's ratio options exist. This limits the shape changing capabilities of the lattice, requiring external shearing forces to generate horizontal deformation. Having no positive Poisson's ratio cell deformation also limits the ability of the lattice to vary cell type in multiple directions.

For the double arrowhead lattice (S3.A), state configurations are defined by the values of each angle $\theta$ for every unit cell joint. For a lattice with AxB joints, there are $2^{A*B}$ potential joint combinations, but the number of valid configurations is greatly limited by geometric restrictions. Having no shearing configuration, unit cell type can be adjusted in stripes, but like the chiral lattice, local cell changes in two directions is limited. To maintain valid physical linkage configurations in the lattice, all cells within a row must maintain a constant value of $\theta$. This means that cell type in the lattice can only be adjusted column by column, reducing the number of total valid combinations from $2^{A*B}$ to $2^B$. With these physical restrictions, we can adjust the effective global Poisson's ratio of the double arrowhead lattice, but we cannot generate spatially varying Poisson's ratios or complex profiles. It should be noted, these restrictions exist with the assumption that the lattice structure remains in the 2D plane. Out of plane deformations may open an even broader design space for deforming lattice structures with star graph representations.

Compared to the chiral and double arrowhead lattice structures, the reentrant honeycomb (S3.B) has a much larger valid combinations space. For a 3x3 grid of reentrant honeycomb cells, there exists a total of 6561 valid states that meet all geometric requirements. We determined the total number of valid reentrant honeycomb states using the validity check algorithm outline in the body of the paper, and S12. To select specific state changes, we use the profile approximation algorithm and the cell assigning algorithms detailed in the body of the paper and S13, S14.

**Required Number of Programming Elements**

A single cell of the inverse honeycomb consists of four vertical linkages and two horizontal linkages. As we compress a lattice, each vertical linkage will flex left or right, making each cell have 4 degrees of freedom. As we expand the cell count to an $NxM$ cell lattice, with N rows and M columns, every cell in the first column (M=1) increments the total linkage count by 4. For multiple columns (M>1), each adjacent cell shares 2 vertical linkages. Accordingly, for every cell in the preceding columns, the total vertical linkage count is incremented by 2. As a result, a lattice has a total of linkage count of $2N*(M+1)$. The biasing mechanisms in our system each constrain the movement of two linkages at a time. Hence, each linkage in the system can be fully constrained by placing actuators at every other element, so the total number of actuators for an $NxM$ cell system is $N*(M+1)$.

**Lattice Profile Generation and Curve Fitting**

We approximated arbitrary functions with corresponding lattice configurations using the algorithms outlined in S14. We did this by first generating evenly spaced x,y points for the desired function. Second, we fit the closest possible mapping for a lattice edge (S13) made up of uniform positive and negatively sloped line segments. Finally, we used the cell generation algorithm (S14) to fill in a valid lattice configuration that will map from the function profile to a flat edge (1,0,1,0,1,0...). Two examples of function approximation can be seen in figures S11.A and S11.B.

Maximum horizontal displacement and maximum possible slope of a lattice become important parameters when selecting the geometry of a lattice and approximating profiles. As shown in figure S1.A, each cell's geometry can be described with horizontal link length $S_1$ and vertical link length $S_2$. The final compressed height of a cell can be expressed as $l_2 = 2S_2 - c$ such that c is the magnitude of the cell's vertical compression and $S_2$ is the vertical link length. If the initial angle of a joint is constrained to $\theta < \pi/2$, the compressed joint angle $\theta = \arcsin\left(\frac{l_2}{2S_1}\right)$ throughout the trajectory of the cell's compression. If the initial angle $\theta > \pi/2$, the compressed angle $\theta = \pi - \arcsin\left(\frac{l_2}{2S_1}\right)$ as the cell compresses. $\theta$ can then be used to find the horizontal displacement of each linkage, $a = S_2\cos(\theta)$. The total effective width of a compressed cell is $l_1 = 2S_1 + 2a$.

**Electrically Actuated Structures**

We fabricated a planar honeycomb structure with small Mgaxyff analog micro linear servo motors to bias the system. The setup included 12 small linear servo motors driven by chained printed circuit boards (PCBs) and controlled by an Arduino. Each PCB consisted of a PCA9685 16 channel servo driver, along with a 4 RGBLED lights and 4 solder jumpers to assign an address for each unit. We fabricated compliant joints from 3D printed TPU filament and all other lattice components from PLA. To program each joint configuration, servo motors slid small plates with cantilevered posts to push each joint to $\frac{\pi}{2} > \theta \ or \ \theta < \frac{\pi}{2}$ (S10). We actuated global transformations in the structure using a single Nema 17 stepper motor connected to a ball screw linear actuator.

To demonstrate feasibility, we programmed the system to transition between four preset states. Figure S10 shows the lattice states as it transitions between the central node and a net positive Poisson's ratio state, a net negative Poisson's ratio state, and a net zero Poisson's ratio state. The structure performed these transformations by first, moving all servo motors to a neutral and unbiased state. Second, using the stepper motor to extend the lattice to its maximum length (central node). Third, actuating all linear servos to bias all joints in the lattice. And finally, using the stepper motor to compress the structure to the new configuration.

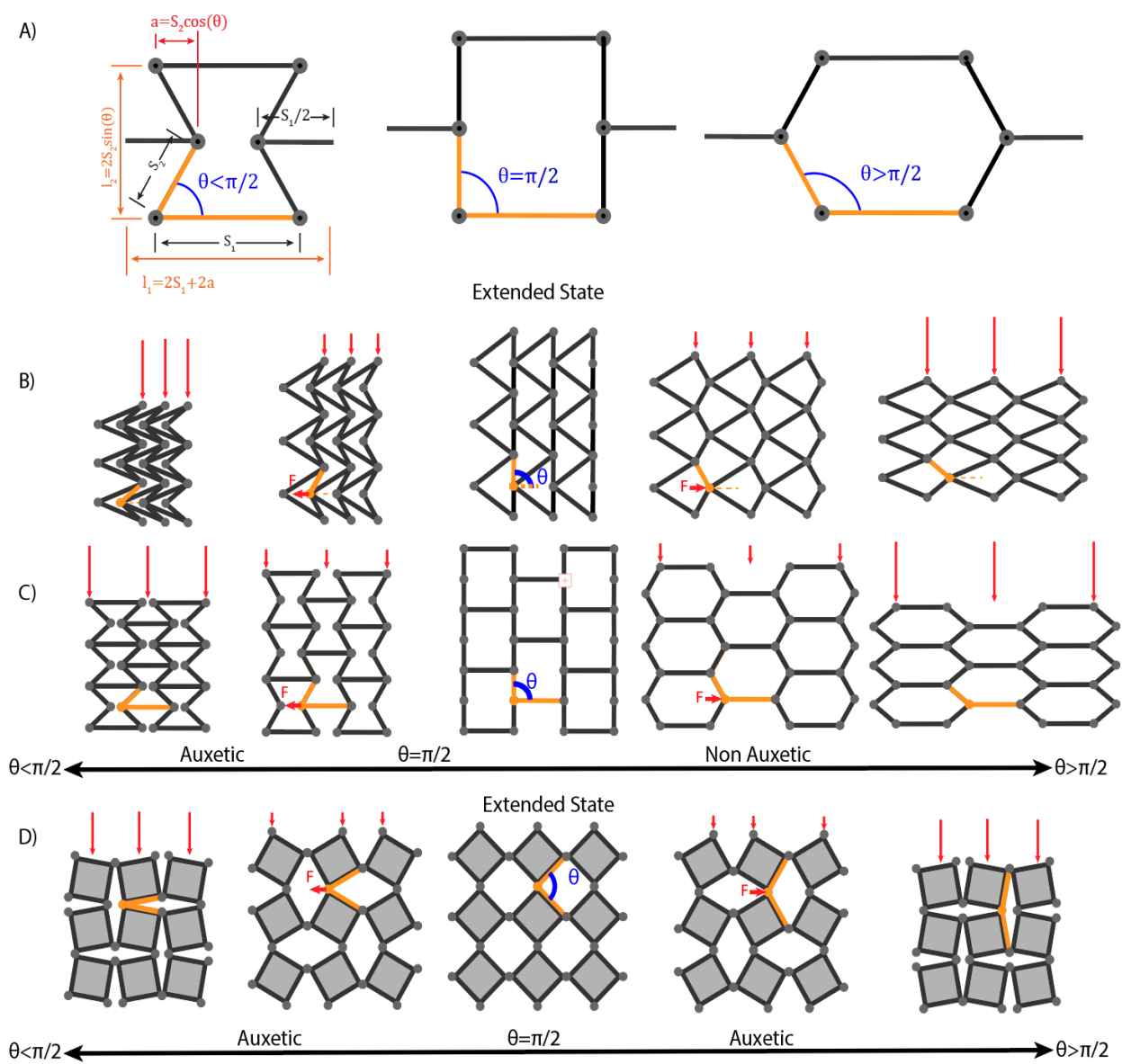

**S1. Transition modes for auxetic lattice structures. A)** We define honeycomb unit cells based on dimensions $S_1$ and $S_2$. The tiling vectors $l_1$ and $l_2$ depend on the dimensions and the angle $\theta$. **B)** The double arrowhead lattice transitions from auxetic to non-auxetic. **B)** The honeycomb structure transitions from auxetic to non-auxetic. **C)** The rotating squares lattice remains auxetic throughout the full range of $\theta$.

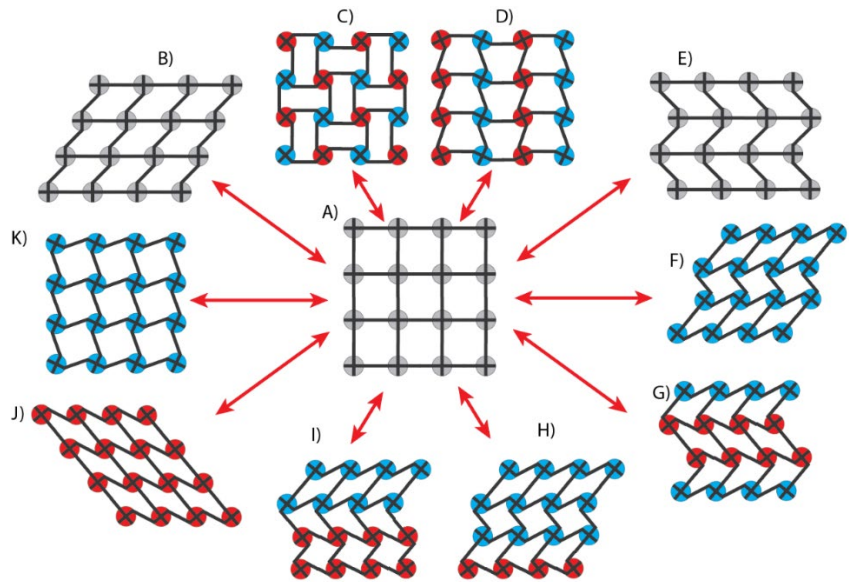

**S2. Star Graph Configuration of Chiral Structures. A)** Undeformed mode with no rotations or shearing is the central transition state for chiral structures. The following states can all be accessed by applying specific loading to individual units within the structure. Red units rotate counterclockwise, blue units rotate clockwise. **B)** Shearing right, no rotations **C)** Checkerboarded CCW, CW rotations **D)** Alternating strips of rotations (CW,CCW) **E)** Shearing left/right, no rotations **F-I)** Example combinations of shearing and rotation **J)** Shearing left, CCW rotations **K)** All CCW rotations.

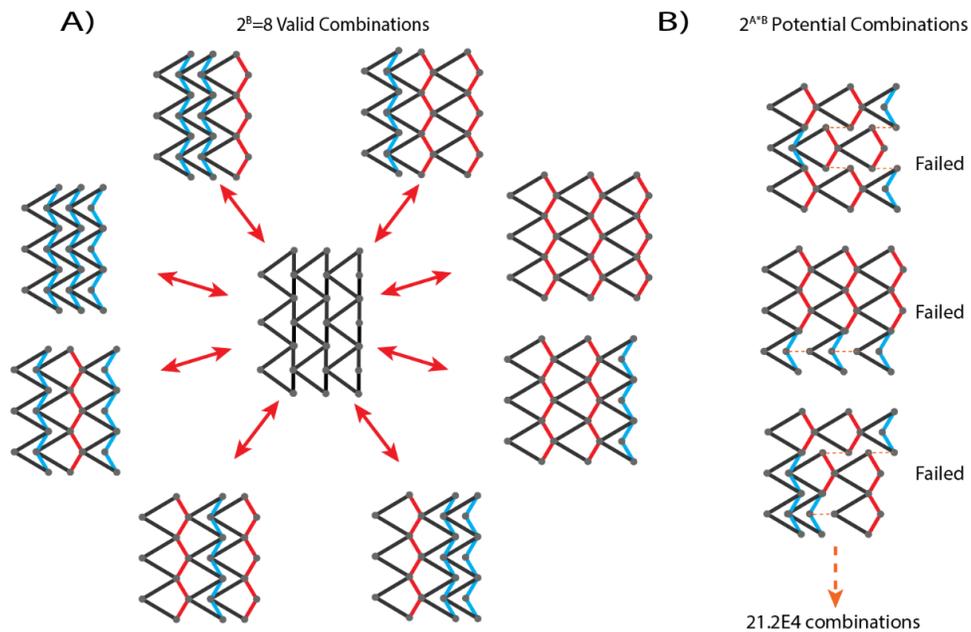

**S3. Star Graph Configuration for A Double Arrowhead Lattice. A)** An AxB double arrowhead lattice with A=6 and B=3 linkages can transition from the central node to $2^B = 8$ different states. **B)** With all positive and negative joint combinations, $2^{6*3} = 21.2E4$ potential linkage variations exist. Dashed orange lines represent discontinuities resulting in linkage deformation.

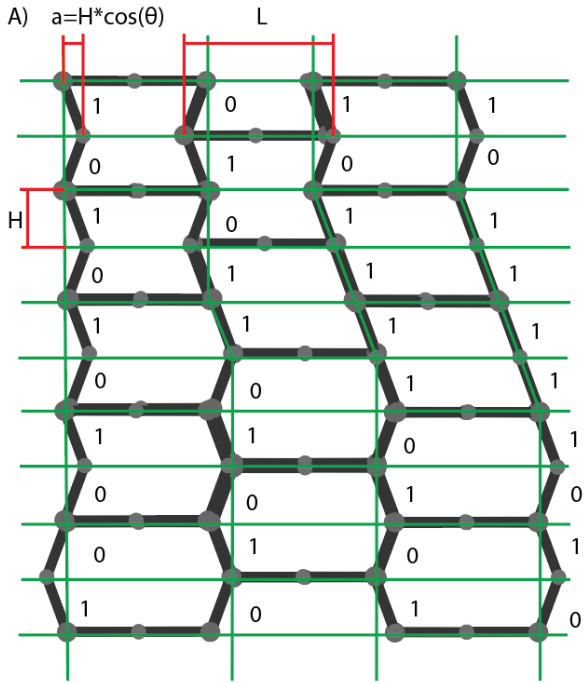

**S4. Example of Valid Structure Combination Check. A)** We encode a structure's vertical linkage slope as either a 1 or a 0. Horizontal linkages span alternating joints. **B)** We use the horizonal joint position to verify that the lattice maintains all horizontal linkage lengths (L).

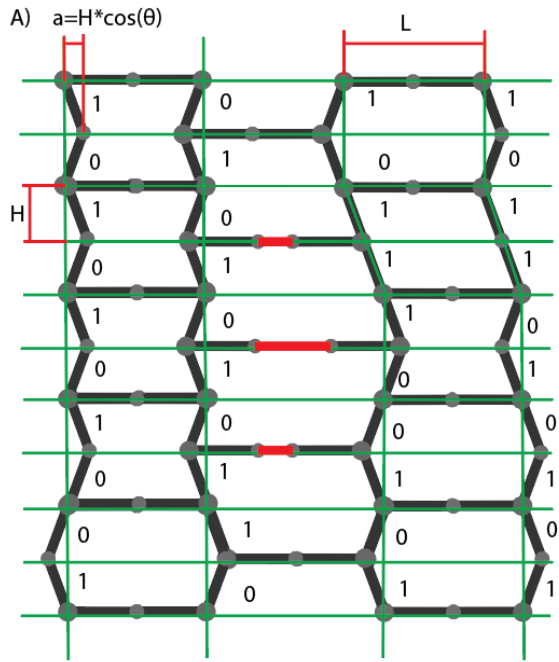

**S5. Example of Invalid Structure Combination Check. A)** Red lines signify invalid linkage lengths. **B)** The distance between joints fails to equal the original linkage length (L) for three horizontal linkages, invalidating the configuration.

| Experimental Total Valid Combinations with A*B<50 | | | | | | | | |
|---|---|---|---|---|---|---|---|---|
| AxB Linkages | 2 | 4 | 6 | 8 | 10 | 12 | 14 | 16 |
| 2 | 6 | 36 | 216 | 1296 | 7776 | 46656 | 279936 | |
| 4 | 36 | 486 | 6642 | 90882 | 1243674 | 17019234 | | |
| 6 | 216 | 6642 | 210924 | 6730128 | | | | |
| 8 | 1296 | 90882 | 6730128 | 475487113 | | | | |
| 10 | 7776 | 1243674 | | | 5.62438E+12 | | | |
| 12 | 46656 | 17019234 | | | | 3.49006E+17 | | |
| 14 | 279936 | | | | | | | |
| | | | fit data | validation data | extrapolated values | | | |

| Total Potential Combinations = 2^(A*B) | | | | | | | | |
|---|---|---|---|---|---|---|---|---|
| AxB Linkages | 2 | 4 | 6 | 8 | 10 | 12 | 14 | 16 |
| 2 | 16 | 256 | 4096 | 65536 | 1048576 | 16777216 | 268435456 | 4.3E+09 |
| 4 | 256 | 65536 | 16777216 | 4294967296 | 1.09951E+12 | 2.81475E+14 | 7.20576E+16 | 1.8E+19 |
| 6 | 4096 | 16777216 | 68719476736 | 2.81475E+14 | 1.15292E+18 | 4.72237E+21 | 1.93428E+25 | 7.9E+28 |
| 8 | 65536 | 4294967296 | 2.81475E+14 | 1.84467E+19 | 1.20893E+24 | 7.92282E+28 | 5.1923E+33 | 3.4E+38 |
| 10 | 1048576 | 1.09951E+12 | 1.15292E+18 | 1.20893E+24 | 1.26765E+30 | 1.32923E+36 | 1.3938E+42 | 1.5E+48 |
| 12 | 16777216 | 2.81475E+14 | 4.72237E+21 | 7.92282E+28 | 1.32923E+36 | 2.23007E+43 | 3.74144E+50 | 6.3E+57 |
| 14 | 268435456 | 7.20576E+16 | 1.93428E+25 | 5.1923E+33 | 1.3938E+42 | 3.74144E+50 | 1.00434E+59 | 2.7E+67 |

| Expression for total valid cells for an AxB lattice | | K1: Slope change with A | 0.2989 |
|---|---|---|---|
| general expression: | $2^{(K_1 AB + K_2 (A+B) + K_3)}$ | K2: Intercept change with A | 0.6924 |
| | | K3: intercept of the intercept | -1.383169 |

| Validation: | | | | | | | | |
|---|---|---|---|---|---|---|---|---|
| A= | 10 | 12 | 14 | 10 | 12 | 6 | 4 | 2 |
| B= | 2 | 2 | 2 | 4 | 4 | 8 | 12 | 14 |
| calculated count= | 7663 | 45831 | 274117 | 1261257 | 17277977 | 6616473 | 17277977 | 274117 |
| experimental counts= | 7776 | 46656 | 279936 | 1243674 | 17019234 | 6730128 | 17019234 | 279936 |
| % error= | 1.45% | 1.77% | 2.08% | 1.41% | 1.52% | 1.69% | 1.52% | 2.08% |

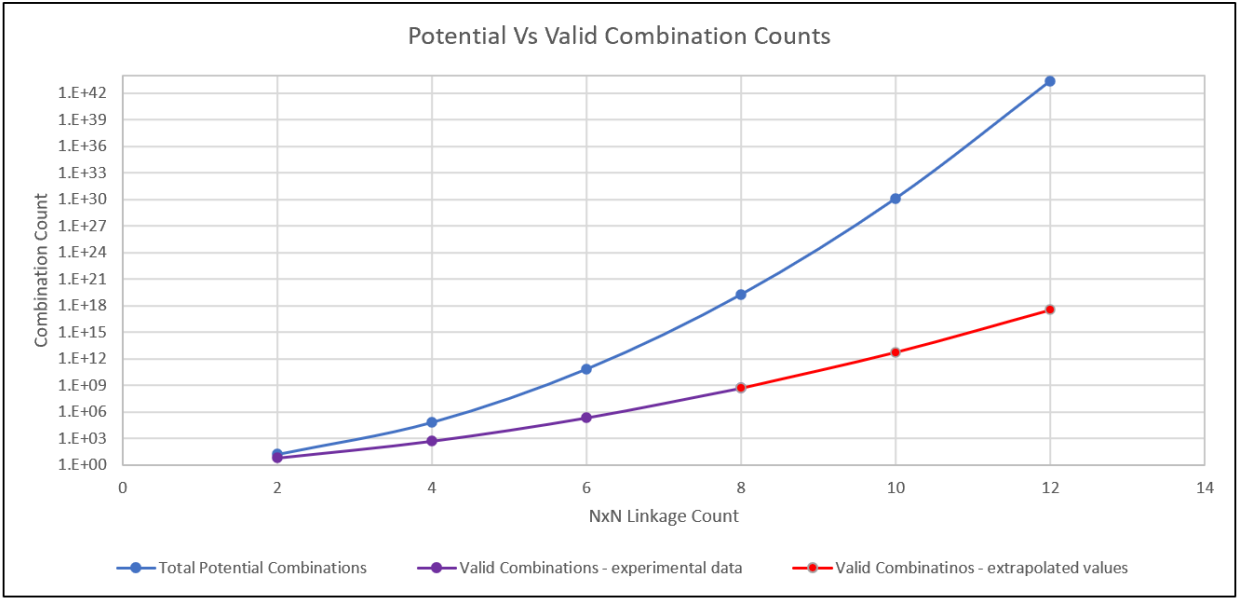

**S6. Combination space for total joint combinations and valid joint combinations.** As lattice cell count increases, total joint combinations expand far more rapidly than valid cell count. Tables A and B display the valid combination count and the total potential combination count in relation to $A \times B$ linkage number lattice dimensions.

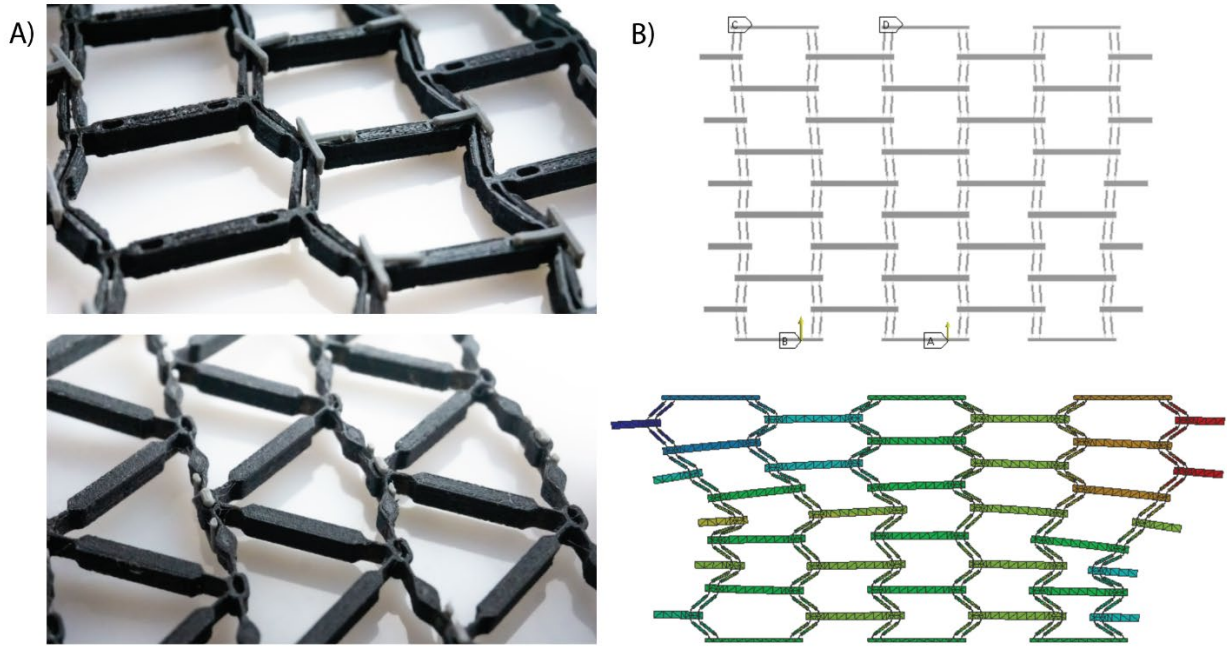

**S7. Fabrication of planar structures. A)** Detailed view of printed honeycomb (top) and arrowhead (bottom) TPU structures with PLA inserts to bias the buckling of the structure. **B)** We used finite element analysis (FEA) simulation to validate our design strategy for creating predicably compressible lattice structures. The pictured structure accurately represents the compression of the physical lattice show in Fig.1.D.

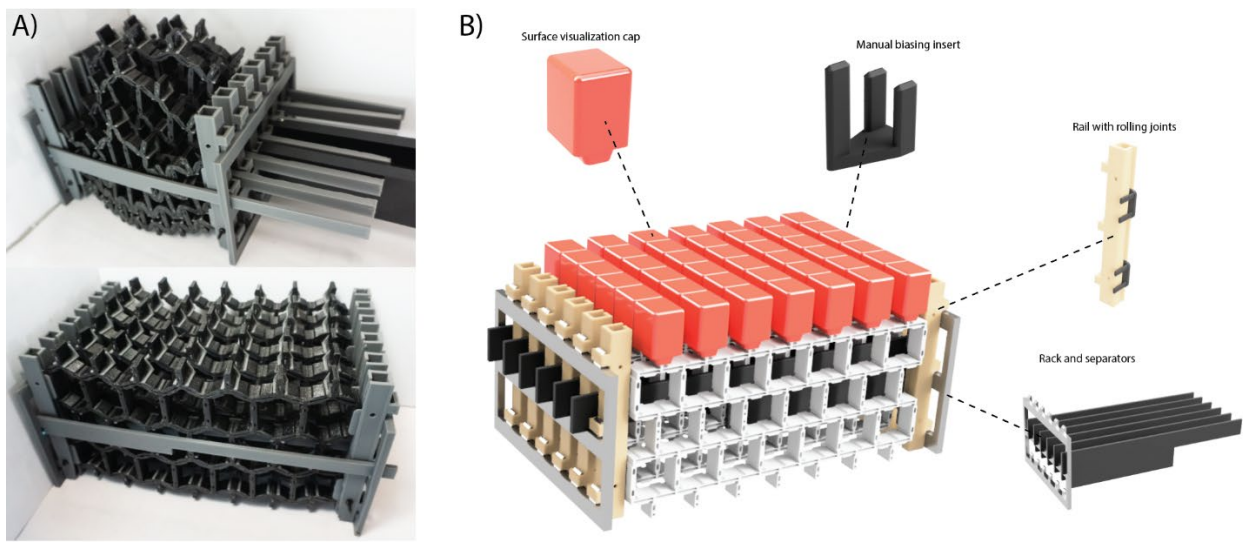

**S8. Binary Surface Reprogramming. A)** 3D structure without caps for visualization. (Top) Structure is configured to compresses to a convex surface**.** (Bottom) Expanded structure creates a uniform flat surface. **B)** Complete assembly of reprogrammable structure is made up of several components.

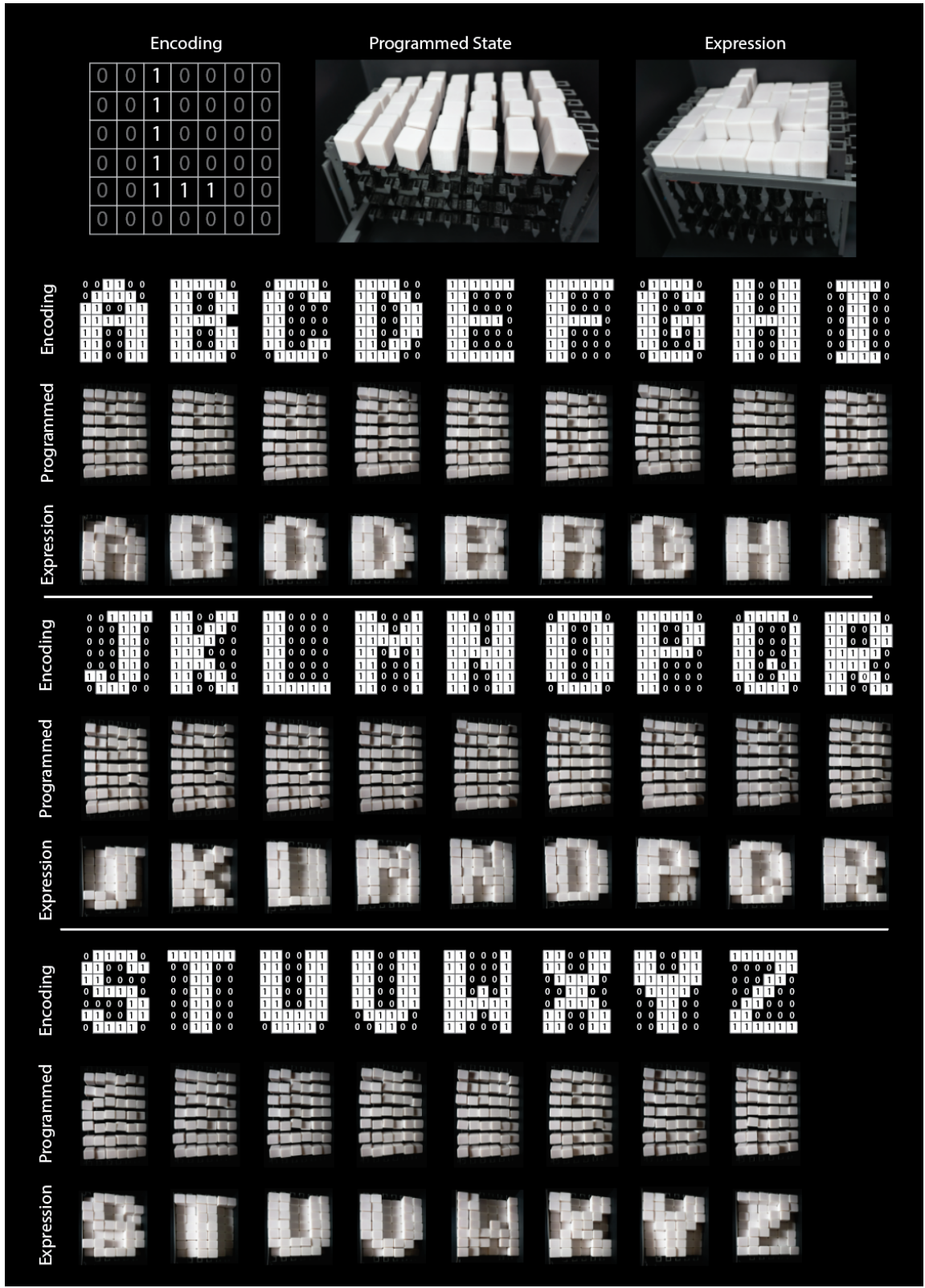

**S9. Binary Surface Reprogramming.** 3D reprogrammable surface structure transitions to display every letter in the English alphabet.

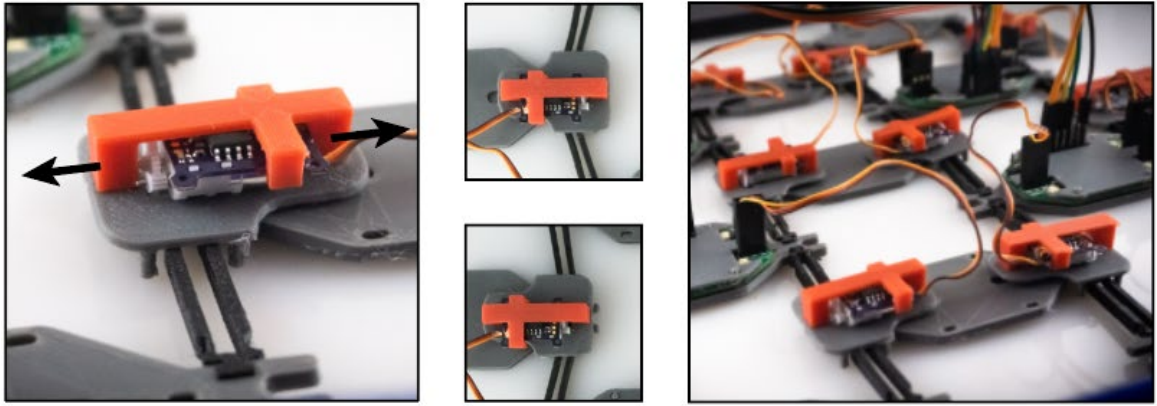

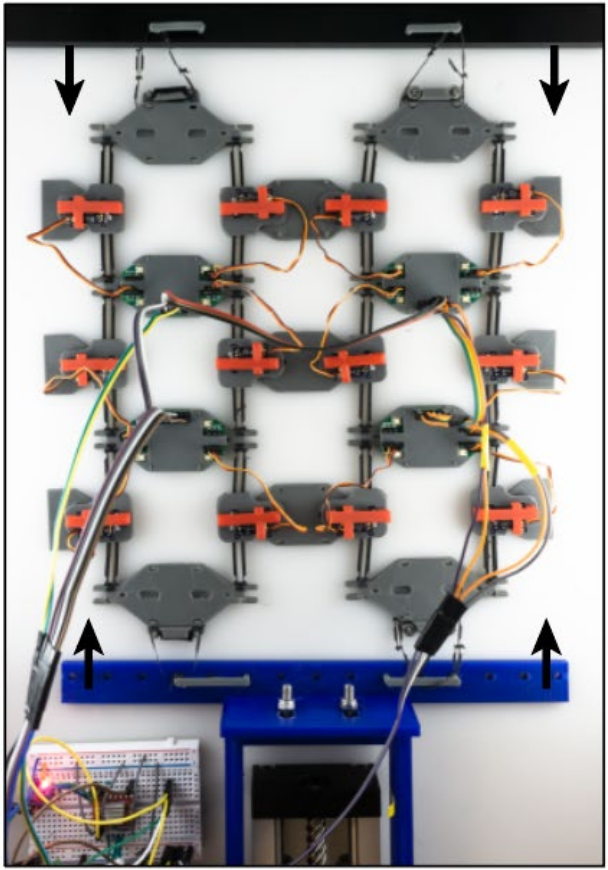

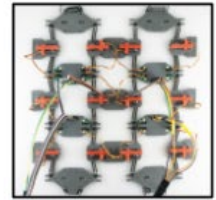
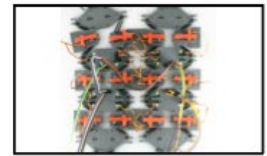

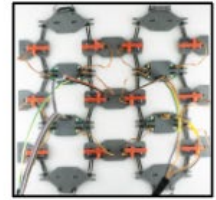
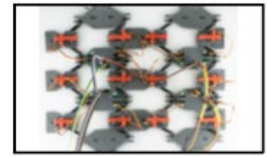

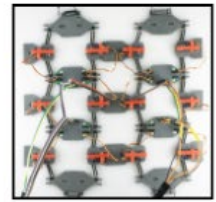
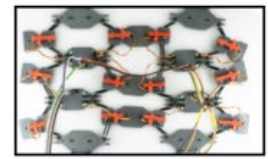

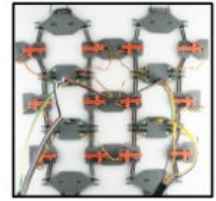
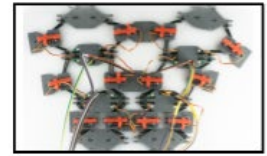

**S10. Robotically Switching Lattice. A)** A stepper motor applied a global compressive force to the structure and individual joints programmed using an array of linear servo motors. **B, C, D)** Lattice is programmed to buckle into multiple different configurations. Top to bottom: Global negative Poisson's ratio, global zero Poisson's ratio, global positive Poisson's ratio, vertical transition in Poisson's ratio from positive to negative.

A) $x = .1((y-7)^2 + 3y + 4)$

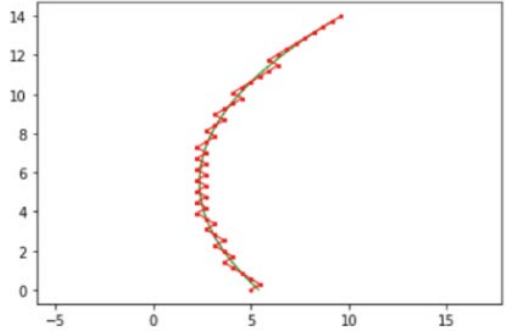
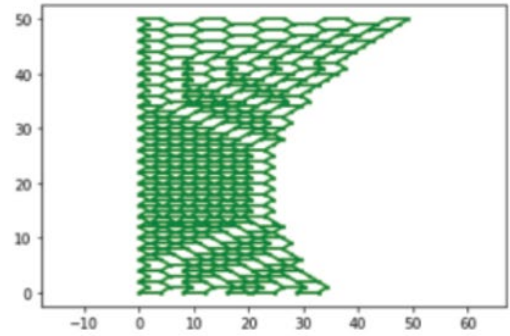

B) $x = -.1((y-4)^2 + 3y + 9)$

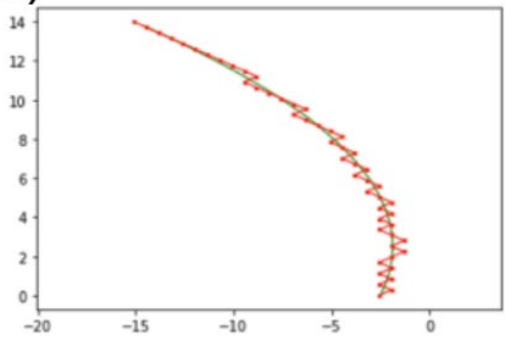
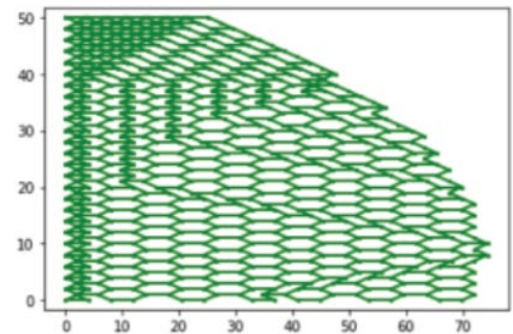

**S11. Function Approximation and Lattice Representation.** Example of concave (A) and convex (B) function estimation using our profile approximation algorithm (S15) and our lattice generation algorithm (S14).

# Pseudo Code

## S12. Pseudo code for validity check:
'Set x values for every joint in the lattice. Check that x values match linkage constraints'

INIT top_left_xval to 0
FOR top_cells in lattice_width
    SET xval to left_xval plus width
    IF top_cell=even
        ADD left_a_values to xval

For all_cells in column
    For all columns
        SET xval to above_xval plus above_a_val

For all_cells in column
    For all columns
        IF column and row number = odd
            IF xval not = left_xval plus width
                Lattice FAILS
        IF column and row number = even
            IF xval not = left_xval plus width
                Lattice FAILS
        ELSE Lattice VALID

## S13. Pseudo code for solving arbitrary profile to flat back:
'Solve joint values for a lattice starting with an arbitrary profile (1,1,0,1,0,0,0..) and ending with a flat back (1,0,1,0,1,0...)'

INIT initial_profile
WHILE not flat_back
    Add 1 to layer_count
    'Define horizontal linkages in lattice'
    IF layer_count = even
        INIT even_horizontal_linkages
    ELSE
        INIT odd_horizontal_linkages

    INIT new_profile
    FOR linkages in initial_profile
        INIT profile_mid_xvalue
        IF linkage_index = horizontal_linkage
            'If 2 links have the same value in between horizontal linkages
            Then the cell is a shear element and fully defined'
            IF linkage_value=next_linkage_value
                SET new_profile_link to linkage_value
                SET next_new_profile_link to next_linkage_value
        ELSE
            'Make x values stabilize towards a center value to flatten profile'
            IF linkage_xval>profile_mid_xvalue
                SET new_profile_link to negative_slope
                SET next_new_profile_link to positive_slope
                'Profile with positive Poisson's ratio'
            ELSE
                SET new_profile_link to positive_slope
                SET next_new_profile_link to negative_slope

'Profile with negative Poisson's ratio'

    SET initial_profile to new_profile
    Call: validity_check
    Call: flat_back_check

RETURN full_lattice

### S14. Pseudo code for profile approximation:

'Take arbitrary function or profile and generate approximation in terms of positive and negative sloped line segments.'

INIT ideal_profile
INIT profile_match

'Slice the ideal profile into x and y points'
FOR number_of_slices in ideal_profile
    INIT profile_x_values
    INIT profile_y_values

'Determine magnitude of lattice compression'
Call: find_max_slope
Call: find_total_compression_distance

'Iterate along the profile assigning positive or negative slopes to the profile approximation'
INIT current_x_value to 0
FOR cells in number of slices
    IF profile_x_value>current_x_value
        ADD positive_value to profile_match
    ELSE
        ADD negative_value to profile_match
RETURN profile_match